\documentclass[10pt,twocolumn,letterpaper]{article}

\usepackage{iccv}
\usepackage{times}
\usepackage{epsfig}
\usepackage{graphicx}
\usepackage{amsmath}
\usepackage{amssymb}
\usepackage{multirow}
\usepackage{array}
\usepackage{bm}

\usepackage[colorlinks, breaklinks=true,bookmarks=false]{hyperref}

\iccvfinalcopy 

\def\iccvPaperID{4701} 

\ificcvfinal\fi

\newcommand*{\affmark}[1][*]{\textsuperscript{#1}}
\newcommand\equalcontribution{\thanks{Contributed equally and work done while Wen Liu was a Research Intern with Tencent AI Lab.}}

\begin{document}
	
	\title{Liquid Warping GAN: A Unified Framework for Human Motion Imitation, Appearance Transfer and Novel View Synthesis}
	
	\author{
		Wen Liu\affmark[1]\equalcontribution \quad\hfill Zhixin Piao\affmark[1]\footnotemark[1] \quad\hfill Jie Min\affmark[1] \quad\hfill  Wenhan Luo\affmark[2] \quad\hfill Lin Ma\affmark[2] \quad\hfill Shenghua Gao\affmark[1]\\
		\affmark[1]ShanghaiTech University \quad  \affmark[2]Tencent AI Lab \\
		{\tt\small \{liuwen,piaozhx,minjie,gaoshh\}@shanghaitech.edu.cn} \\ 
		{\tt\small \{whluo.china,forest.linma\}@gmail.com}
	}
	
	\maketitle
	\thispagestyle{empty}

	\begin{abstract}
		We tackle the human motion imitation, appearance transfer, and novel view synthesis within a unified framework, which means that the model once being trained can be used to handle all these tasks. The existing task-specific methods mainly use 2D keypoints (pose) to estimate the human body structure. However, they only expresses the position information with no abilities to characterize the personalized shape of the individual person and model the limbs rotations. In this paper, we propose to use a 3D body mesh recovery module to disentangle the pose and shape, which can not only model the joint location and rotation but also characterize the personalized body shape. To preserve the source information, such as texture, style, color, and face identity, we propose a Liquid Warping GAN with Liquid Warping Block (LWB) that propagates the source information in both image and feature spaces, and synthesizes an image with respect to the reference. Specifically, the source features are extracted by a denoising convolutional auto-encoder for characterizing the source identity well. Furthermore, our proposed method is able to support a more flexible warping from multiple sources. In addition, we build a new dataset, namely Impersonator (iPER) dataset, for the evaluation of human motion imitation, appearance transfer, and novel view synthesis. Extensive experiments demonstrate the effectiveness of our method in several aspects, such as robustness in occlusion case and preserving face identity, shape consistency and clothes details. All codes and datasets are available on \url{https://svip-lab.github.io/project/impersonator.html}.
	\end{abstract}
	
	\vspace{-5mm}
	\section{Introduction}
\begin{figure}[t]
		\begin{center}
			\includegraphics[width=\linewidth]{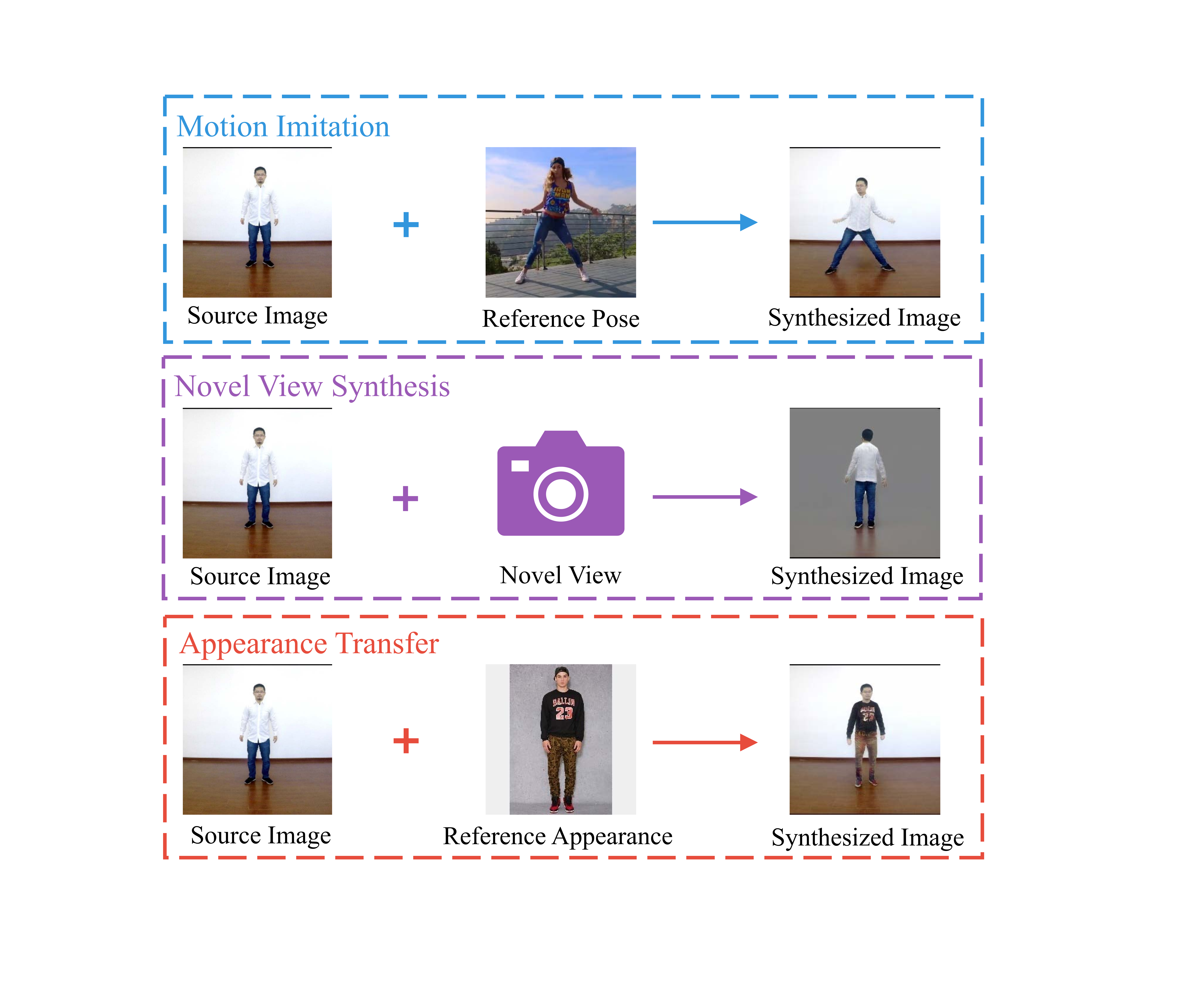}
		\end{center}
		\vspace{-3mm}
		\caption{Illustration of human motion imitation, appearance transfer and novel view synthesis. The first column is the source image and the second column is reference condition, such as image or novel view of camera. The third column is the synthesized results.}
		\label{fig:examples}
		\vspace{-2mm}
	\end{figure}

	Human image synthesis, including human motion imitation~\cite{posewarp2018,pG2017nips,DSC2018}, appearance transfer~\cite{swapnet2018,HAT_2018_CVPR} and novel view synthesis~\cite{Zhao0C0JF18,Zhu_2018_CVPR}, has huge potential applications in re-enactment, character animation, virtual clothes try-on, movie or game making and so on. The definition is that given a source human image and a reference human image, i) the goal of motion imitation is to generate an image with texture from source human and pose from reference human, as depicted in the top of Fig.~\ref{fig:examples}; ii) human novel view synthesis aims to synthesize new images of the human body, captured from different viewpoints, as illustrated in the middle of Fig.~\ref{fig:examples}; iii) the goal of appearance transfer is to generate a human image preserving reference identity with clothes, as shown in the bottom of Fig.~\ref{fig:examples} where different parts might come from different people.
	
\begin{figure}[t]
		\centering
		\includegraphics[width=\linewidth]{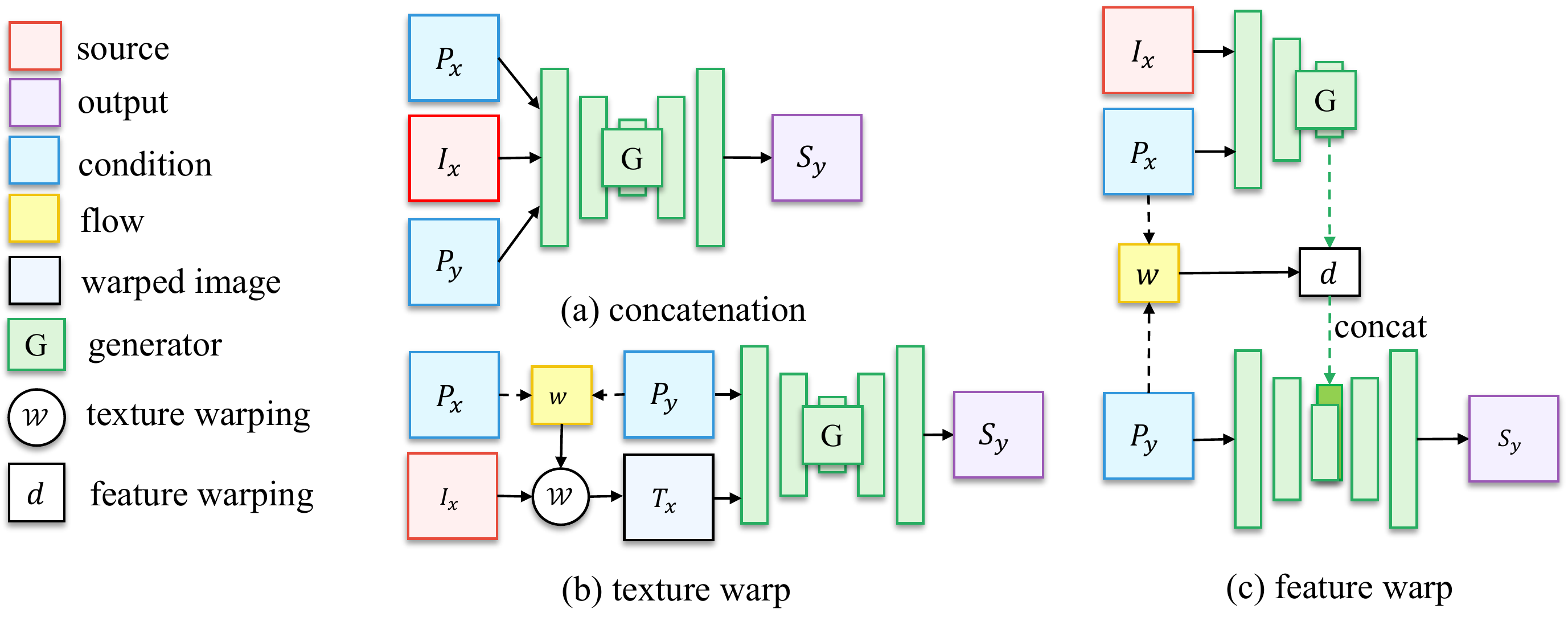}
		\caption{Three existing approaches of propagating source information into target condition. (a) is early concatenation, and it concatenates the source image and source condition, as well as target condition, into the color channel. (b) and (c) are texture and feature warping, respectively, and the source image or its features are propagated into target condition under a fitted transformation flow.}
		\label{fig:fusion} 
		\vspace{-2mm}
	\end{figure}

	In the realm of human image synthesis, previous works separately handle these tasks~\cite{pG2017nips,swapnet2018,Zhu_2018_CVPR} with task-specific pipeline, which seems to be difficult to extend to other tasks.
	Recently, generative adversarial network (GAN)~\cite{gan2014} achieves great successes on these tasks. Taking human motion imitation as an example, we summarize recent approaches in Fig.~\ref{fig:fusion}. In an early work~\cite{pG2017nips}, as shown in Fig.~\ref{fig:fusion} (a), source image (with its pose condition) and target pose condition are concatenated  which thereafter is fed into a network with adversarial training to generate an image with desired pose. However, direct concatenation does not take the spatial layout into consideration, and it is ambiguous for the generator to place the pixel from source image into a right position. Thus, it always results in a blurred image and loses the source identity. Later, inspired by the spatial transformer networks (STN)~\cite{STN2015}, a texture warping method~\cite{posewarp2018}, as shown in Fig.~\ref{fig:fusion} (b), is proposed. It firstly fits a rough affine transformation matrix from source and reference poses, uses an STN to warp the source image into reference pose and generates the final result based on the warped image. Texture warping, however, could not preserve the source information as well, in terms of the color, style or face identity, because the generator might drop out source information after several down-sampling operations,  such as stride convolution and pooling. Meanwhile, contemporary works~\cite{softgate18,DSC2018} propose to warp the deep features of the source images into target pose rather than that in image space, as shown in Fig~\ref{fig:fusion} (c), named as feature warping.  However, features extracted by an encoder in feature warping cannot guarantee to accurately characterize the source identity and thus consequently produce a blur or low-fidelity image in an inevitable way. 
	
	The aforementioned existing methods encounter with challenges in generating unrealistic-looking images, due to three reasons: 1) diverse clothes in terms of texture, style, color, and high-structure face identity are difficult to be captured and preserved in their network architecture; 2) articulated and deformable human bodies result in a large spatial layout and geometric changes for arbitrary pose manipulations; 3) all these methods cannot handle multiple source inputs, such as in appearance transfer, different parts might come from different source people.
	
	In this paper, to preserve the source information, including details of clothes and face identity, we propose a Liquid Warping Block (LWB) to address the loss of source information from three aspects: 1) a denoising convolutional auto-encoder is used to extract useful features that preserve source information, including texture, color, style and face identity; 2) source features of each local part are blended into a global feature stream by our proposed LWB to further preserve the source details; 3) it supports multiple-source warping, such as in appearance transfer, warping the features of head from one source and those of body from another, and aggregating into a global feature stream. This will further enhance the local identity of each source part. 
	
	In addition, existing approaches mainly rely on 2D pose~\cite{posewarp2018, pG2017nips, DSC2018}, dense pose~\cite{DensePoseTransfer} and body parsing~\cite{softgate18}. These methods only take care of the layout locations and ignore the personalized shape and limbs (joints) rotations, which are even more essential than layout location in human image synthesis. For example, in an extreme case, a tall man imitates the actions of a short person and using the 2D skeleton, dense pose and body parsing condition will unavoidably change the height and size of the tall one, as shown in the bottom of Fig.~\ref{fig:comparison}. To overcome their shortcomings, we use a parametric statistical human body model, SMPL~\cite{SMPLify, SMPL:2015, HMR} which disentangles human body into pose (joint rotations) and shape. It outputs 3D mesh (without clothes) rather than the layouts of joints and parts. Further, transformation flows can be easily calculated by matching the correspondences between two 3D triangulated meshes, which is more accurate and results in fewer misalignments than previous fitted affine matrix from keypoints~\cite{posewarp2018,DSC2018}. 
	
	Based on SMPL model and Liquid Warping Block (LWB), our method can be further extended into other tasks, including human appearance transfer and novel view synthesis for free and one model can handle these three tasks. We summarize our contributions as follows: 1) we propose a LWB to propagate and address the loss of the source information, such as texture, style, color, and face identity, in both image and feature space; 2) by taking advantages of both LWB and the 3D parametric model, our method is a unified framework for human motion imitation, appearance transfer, and novel view synthesis; 3) we build a dataset for these tasks, especially for human motion imitation in video, and all codes and datasets are released for further research convenience in the community.
	
	\begin{figure*}[t] 
		\centering
		\includegraphics[width=0.9\linewidth]{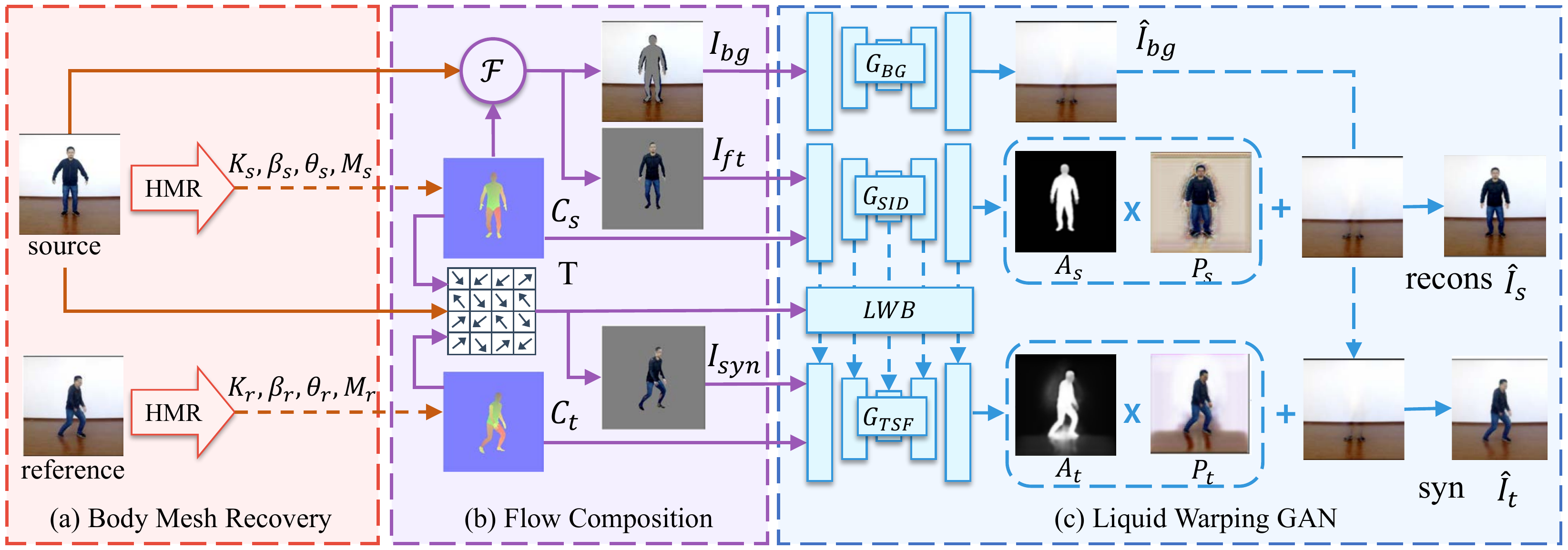}   
		\caption{The training pipeline of our method. We randomly sample a pair of images from a video, denoting one of them as source image, named $I_s$ and the other as reference image named $I_r$. \textbf{(a)} A body mesh recovery module will estimate the 3D mesh of each image, and render their correspondence map, $C_s$ and $C_t$; \textbf{(b)} The flow composition module will first calculate the transformation flow $T$ based on two correspondence maps and their projected vertices in image space. Then it will separate source image $I_s$ into foreground image $I_{ft}$ and masked background $I_{bg}$. Finally it warps the source image based on transformation flow $T$, and produces a warped image $I_{syn}$; \textbf{(c)} In the last GAN module, the generator consists of three streams, which separately generates the background image $\hat{I}_{bg}$ by $G_{BG}$, reconstructs the source image $\hat{I}_s$ by $G_{SID}$ and synthesizes the target image $\hat{I}_t$ under reference condition by $G_{TSF}$. To preserve the details of source image, we propose a novel Liquid Warping Block (LWB, shown in Fig.~\ref{fig:lwb}) which propagates the source features of $G_{SID}$ into $G_{TSF}$ at several layers and preserve the source information, in terms of texture, style and color.}
		\label{fig:pipeline} 
		\vspace{-3mm}
	\end{figure*}
	
	\section{Related Work}
	\textbf{Human Motion Imitation.} Recently, most methods are based on conditioned generative adversarial networks (CGAN)~\cite{posewarp2018,chan2018everybody,pG2017nips, ma2018disentangled,DensePoseTransfer,Si_2018_CVPR} or Variational Auto-Encoder~\cite{vunet2018}. Their key technical idea is to combine target image along with source pose (2D key-points) as inputs and generate realistic images by GANs using source pose. The difference of those approaches are merely in network architectures and adversarial losses. In~\cite{pG2017nips}, a U-Net generator is designed and a coarse-to-fine strategy is utilized to generate $256\times256$ images. 
	Si~\etal~\cite{posewarp2018,Si_2018_CVPR} propose a multistage adversarial loss and separately generate the foreground (or different body parts) and background. Neverova~\etal~\cite{DensePoseTransfer} replace the sparse 2D key-points with the dense correspondences between image and surface of the human body by DensePose~\cite{DensePose}. Chan ~\etal\cite{chan2018everybody} use pix2pixHD~\cite{Wang_2018_CVPR} framework together with a specialized Face GAN to learn a mapping from 2D skeleton to image and generate a more realistic target image. Furthermore, Wang~\etal\cite{wangVID2VID} extend it to video generation and Liu~\etal\cite{Liu2018Neural} propose a neural renderer of human actor video. However, their works just train a mapping from 2D pose (or parts) to image of each person --- in other words, every body need to train their own model. This shortcoming might limit its wide application.
	
	\textbf{Human Appearance Transfer.} Human appearance modeling or transfer is a vast topic, especially in the field of virtual try-on applications, from computer graphics pipelines~\cite{ponsmollSIGGRAPH17clothcap} to learning based pipelines~\cite{swapnet2018,HAT_2018_CVPR}. Graphics based methods first estimate the detailed 3D human mesh with clothes via garments and 3d scanners~\cite{3dScanZhang17} or multiple camera arrays~\cite{mvLeroyFB17} and then human appearance with clothes is capable to be conducted from one person to another based on the detailed 3D mesh. Although these methods can produce high-fidelity result,
	their cost, size and controlled environment are unfriendly and inconvenient to customers. Recently, in the light of deep generative models, SwapNet~\cite{swapnet2018} firstly learns a pose-guided clothing segmentation synthetic network, and then  the clothing parsing results with texture features from source image feed into an encoder-decoder network to generate the image with desired garment. In~\cite{HAT_2018_CVPR}, the authors leverage a geometric 3D shape model combined with learning methods, swap the color of visible vertices of the triangulated mesh and train a model to infer that of invisible vertices.
	
	\textbf{Human Novel View Synthesis.} Novel view synthesis aims to synthesize new images of the same object, as well as the human body, from arbitrary viewpoints. The core step of existing methods is to fit a correspondence map from the observable views to novel views by convolutional neural networks. In~\cite{ZhouTSME16}, the authors use CNNs to predict appearance flow and synthesize new images of the same object by copying the pixel from source image based on the appearance flow, and they have achieved decent results of rigid objects like vehicles. Following work~\cite{Park_2017_CVPR} proposes to infer the invisible textures based on appearance flow and adversarial generative network (GAN)~\cite{gan2014}, while Zhu~\etal~\cite{Zhu_2018_CVPR} argue that appearance flow based method performs poorly on articulated and deformable objects, such as human bodies. They propose an \emph{appearance-shape-flow} strategy for synthesizing novel views of human bodies. Besides, Zhao~\etal~\cite{Zhao0C0JF18} design a GAN based method to synthesize high-resolution views in a coarse-to-fine way. 
	
	\section{Method}
	
	Our Liquid Warping GAN contains three stages, body mesh recovery, flow composition and a GAN module with Liquid Warping Block (LWB). The training pipeline is the same for different tasks. Once the model has been trained on one task, it can deal with other tasks as well. Here, we use motion imitation as an example, as shown in Fig.~\ref{fig:pipeline}. Denoting the source image as $I_s$ and the reference image $I_r$. The first body mesh recovery module will estimate the 3D mesh of $I_s$ and $I_r$, and render their correspondence maps, $C_s$ and $C_t$. Next, the flow composition module will first calculate the transformation flow $T$ based on two correspondence maps and their projected mesh in image space. The source image $I_s$ is thereby decomposed as front image $I_{ft}$ and masked background $I_{bg}$, and warped to $I_{syn}$ based on transformation flow $T$. The last GAN module has a generator with three streams. It separately generates background image by $G_{BG}$, reconstructs the source image $\hat{I}_s$ by $G_{SID}$ and synthesizes the image $\hat{I}_t$ under reference condition by $G_{TSF}$. To preserve the details of source image, we propose a novel Liquid Warping Block (LWB) and it propagates the source features of $G_{SID}$ into $G_{TSF}$ at several layers.
	
	\subsection{Body Mesh Recovery Module}
	As shown in Fig.~\ref{fig:pipeline} (a), given source image $I_s$ and reference image $I_r$, the role of this stage is to predict the kinematic pose (rotation of limbs) and shape parameters, as well as 3D mesh of each image. In this paper, we use the HMR~\cite{HMR} as 3D pose and shape estimator due to its good trade-off between accuracy and efficiency. In HMR, an image is firstly encoded into a feature with $\mathbb{R}^{2048}$ by a ResNet-50~\cite{resnetv2He16} and then followed by an iterative 3D regression network that predicts the pose $\theta \in\mathbb{R}^{72}$ and shape $\beta \in\mathbb{R}^{10}$ of SMPL~\cite{SMPL:2015}, as well as the weak-perspective camera $K\in\mathbb{R}^3$. SMPL is a 3D body model that can be defined as a differentiable function $M(\theta, \beta) \in \mathbb{R}^{N_v \times 3}$, and it parameterizes a triangulated mesh by $N_v = 6,890$ vertices and $N_f = 13,776$ faces with pose parameters $\theta \in\mathbb{R}^{72}$ and $\beta \in\mathbb{R}^{10}$. Here, shape parameters $\beta$ are coefficients of a low-dimensional shape space learned from thousands of registered scans and the pose parameters $\theta$ are the joint rotations that articulate the bones via forward kinematics. With such process, we will obtain the body reconstruction parameters of source image, $\{K_s, \theta_s, \beta_s, M_s\}$ and those of reference image, $\{K_r, \theta_r, \beta_r, M_r\}$, respectively.
	
	\subsection{Flow Composition Module}
	\label{sec:stage2}
	Based on the previous estimations, we first render a correspondence map of source mesh $M_s$ and that of reference mesh $M_r$ under the camera view of $K_s$. Here, we denote the source and reference correspondence maps as $C_s$ and $C_t$, respectively. In this paper, we use a fully differentiable renderer, Neural Mesh Renderer (NMR)~\cite{cvprKatoUH18}. We thereby project vertices of source $V_s$ into 2D image space by weak-perspective camera, $v_s=Proj(V_s, K_s)$. Then, we calculate the barycentric coordinates of each mesh face, and obtain $f_s \in \mathbb{R}^{N_f \times 2} $. Next, we calculate the transformation flow $T\in\mathbb{R}^{H\times W\times 2}$ by matching the correspondences between source correspondence map with its mesh face coordinates $f_s$ and reference correspondence map. Here $H\times W$ is the size of image.
	Consequently, a front image $I_{ft} $ and a masked background image $I_{bg}$ are derived from masking the source image $I_s$ based on $C_s$. Finally, we warp the source image $I_s$ by the transformation flow $T$, and obtain the warped image $I_{syn}$, as depicted in Fig.~\ref{fig:pipeline}.
	
	
	\begin{figure}[t]
		\begin{center}
			\includegraphics[width=\linewidth]{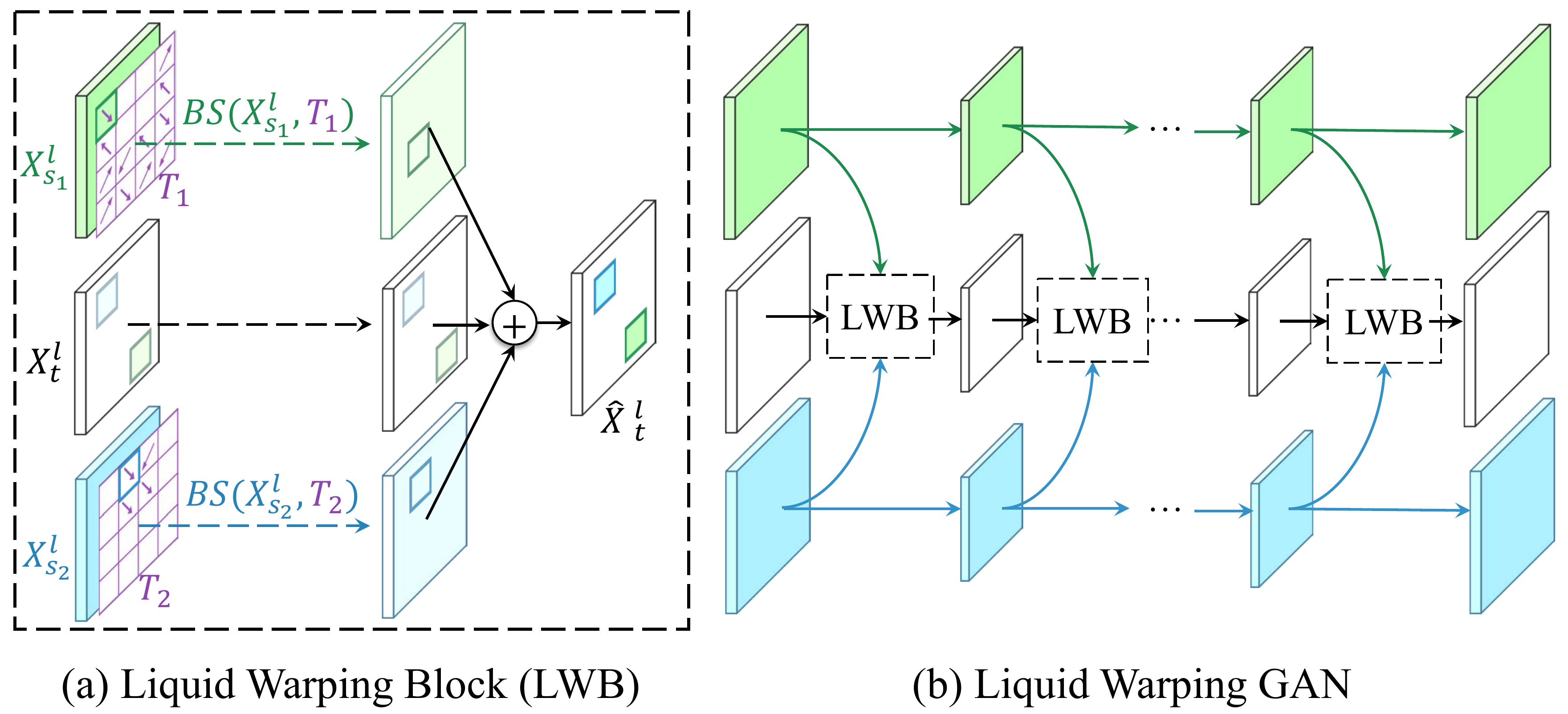}
		\end{center}
		\vspace{-4mm}
		\caption{Illustration of Liquid Warping Block. \textbf{(a)} is the structure of LWB.  $X^{l}_{s_1}$ and $X^{l}_{s_2}$ are the feature maps extracted by $G_{SID}$ of different sources in $l^{th}$ layers. $X^{l}_t$ is the feature map of $G_{TSF}$ at the $l^{th}$ layer. Final output features $\widehat{X}_t^{l}$ aggregate the feature from $G_{TSF}$ and warped source features by bilinear sampler (BS) with respect to the flow $T_1$ and $T_2$. \textbf{(b)} is the architecture of LWB.}
		\label{fig:lwb}
		\vspace{-4mm}
	\end{figure}
	
	\subsection{Liquid Warping GAN}
	This stage synthesizes high-fidelity human image under the desired condition. More specifically, it 1) synthesizes the background image; 2) predicts the color of invisible parts based on the visible parts; 3) generates pixels of clothes, hairs and others out of the reconstruction of SMPL. 
	
	\textbf{Generator.} Our generator works in a three-stream manner. One stream, named $G_{BG}$,
	works on the concatenation of the masked background image $I_{bg}$ and the mask obtained by the binarization of $C_s$ in color channel (4 channels in total) to generate the realistic background image $\hat{I}_{bg}$, as shown in the top stream of Fig.~\ref{fig:pipeline} (c). The other two streams are source identity stream, namely $G_{SID}$ and transfer stream, namely $G_{TSF}$. $G_{SID}$ is a denoising convolutional auto-encoder which aims to guide the encoder to extract the features that are capable to preserve the source information. Together with the $\hat{I}_{bg}$, it takes the masked source foreground $I_{ft}$ and the correspondence map $C_s$ (6 channels in total) as inputs, and reconstructs source front image $\hat{I}_s$. $G_{TSF}$ stream synthesizes the final result , which receives the warped foreground by bilinear sampler and the correspondence map $C_t$ (6 channels in total) as inputs. To preserve the source information, such as texture, style and color, we propose a novel Liquid Warping Block (LWB) that links the source with target streams. It blends the source features from $G_{SID}$ and fuses them into transfer stream $G_{TSF}$, as shown in the bottom of Fig.~\ref{fig:pipeline} (c). 
	
	%
	
	One advantage of our proposed Liquid Warping Block (LWB) is that it addresses multiple sources, such as in human appearance transfer, preserving the head of source one, and wearing the upper outer garment from the source two, while wearing the lower outer garment from the source three. The different parts of features are aggregated into $G_{TSF}$ by their own transformation flow, independently. Here, we take two sources as an example, as shown in Fig.~\ref{fig:lwb}. Denoting $X^{l}_{s_1}$ and $X^{l}_{s_2}$ as the feature maps extracted by $G_{SID}$ of different sources in the $l^{th}$ layer. $X^{l}_t$ is the feature map of $G_{TSF}$ at the $l^{th}$ layer. Each part of source feature is warped by their own transformation flow, and aggregated into the features of $G_{TSF}$. We use bilinear sampler (BS) to warp the source features $X^{l}_{s_1}$ and $X^{l}_{s_2}$, with respect to the transformation flows, $T_1$ and $T_2$, respectively. The final output feature is obtained as follows: 
	\vspace{-2mm}
	\begin{equation*}
	\begin{aligned}
	\widehat{X}_t^{l} &= BS(X^{l}_{s_1}, T_1) + BS(X^{l}_{s_2}, T_2) + X_t^{l}.
	\end{aligned}
	\label{equ:lcb}
	\vspace{-3mm}
	\end{equation*}
	Please note that we only take two sources an example, which can be easily extended to multiple sources.
	
	$G_{BG}$, $G_{SID}$ and $G_{TSF}$ have the similar architecture, named ResUnet, a combination of ResNet~\cite{resnetHe16} and U-Net~\cite{unet2015} without sharing parameters. For $G_{BG}$, we directly regress the final background image, while for $G_{SID}$ and $G_{TSF}$, we concretely generate an attention map $A$ and a color map $P$, as illustrated in Fig.~\ref{fig:pipeline} (c). The final image can be obtained as follows:
	\begin{equation*}
	\vspace{-2mm}
	\begin{aligned}
	\hat{I}_s &= P_s * A_s + \hat{I}_{bg} * (1 - A_s) \\ 
	\hat{I}_t &= P_t * A_t + \hat{I}_{bg} * (1 - A_t).
	\end{aligned}
	\label{equ:att_color}
	\end{equation*}
	\textbf{Discriminator.} For discriminator, we follow the architecture of Pix2Pix~\cite{Isola2017ImagetoImageTW}. More details about our network architectures are provided in supplementary materials.
	
	\subsection{Training Details and Loss Functions}
	In this part, we will introduce the loss functions, and how to train the whole system. For body recovery module, we follow the network architecture and loss functions of HMR~\cite{HMR}. Here, we use a pre-trained model of HMR.
	
	For the Liquid Warping GAN, in the training phase, we randomly sample a pair of images from each video and set one of them as source $I_s$, and another as reference $I_r$. Note that our proposed method is a unified framework for motion imitation, appearance transfer and novel view synthesis. Therefore once the model has been trained, it is capable to be applied to other tasks and does not need to train from scratch. In our experiments, we train a model for motion imitation and then apply it to other tasks, including appearance transfer and novel view synthesis.
	
	The whole loss function contains four terms and they are perceptual loss~\cite{eccvJohnsonAF16}, face identity loss, attention regularization loss and adversarial loss.
	
	\begin{figure*}[t]
		\begin{center}
			\includegraphics[width=\linewidth]{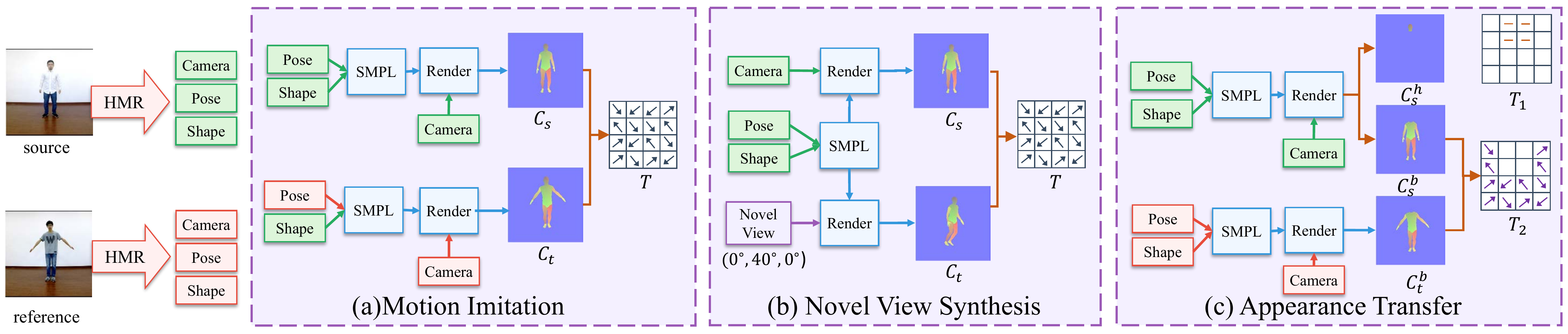}
		\end{center}
		\vspace{-5mm}
		\caption{Illustration of calculating the transformation flows of different tasks during the testing phase. The left is the disentangled body parameters by Body Recovery module of both source and reference images. The right is the different implementations to calculate the transformation flow in different tasks.}
		\vspace{-2mm}
		\label{fig:inference}
	\end{figure*}
	
	\begin{figure*}[t]
		\centering
		\includegraphics[width=\linewidth]{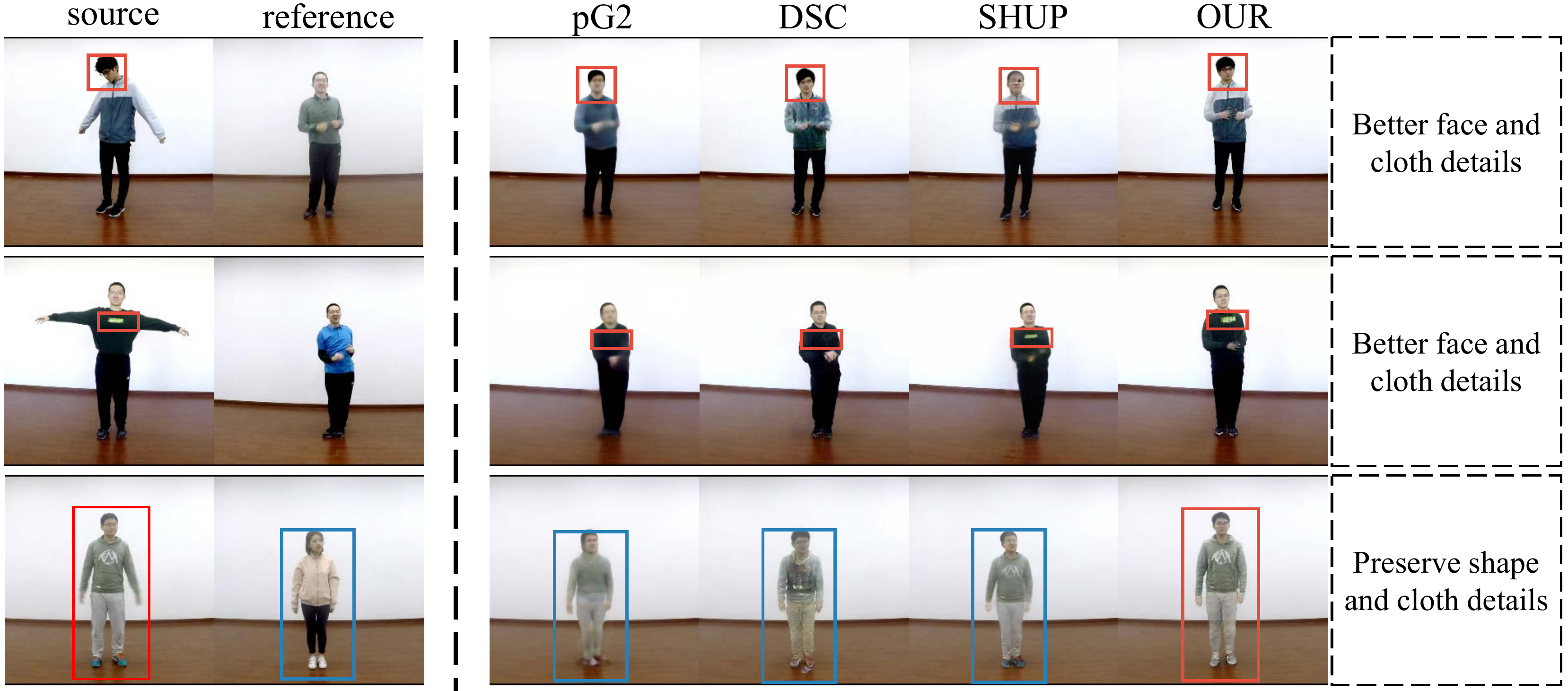}
		\caption{Comparison of our method with others of motion imitation on the iPER dataset (zoom-in for the best of view). 2D pose-guided methods~pG2~\cite{pG2017nips}, DSC~\cite{DSC2018} and SHUP~\cite{posewarp2018} cannot preserve the clothes details, face identity and shape consistency of source images. We highlight the details by red and blue rectangles.}
		\vspace{-4mm}
		\label{fig:comparison} 
	\end{figure*}
	
	\textbf{Perceptual Loss.} It regularizes the reconstructed source image $\hat{I}_s$ and generated target image $\hat{I}_t$ to be closer to the ground truth $I_s$ and $I_r$ in VGG~\cite{Simonyan15} subspace. Its formulation is given as follows: 	\vspace{-2mm}
	\begin{equation*}
	\begin{aligned}
	L_p=\|f(\hat{I}_s) - f(I_s)\|_1 + \|f(\hat{I}_t) - f(I_r)\|_1.
	\end{aligned}
	\label{equ:percep}
	\vspace{-2mm}
	\end{equation*}
	Here, $f$ is a pre-trained VGG-19~\cite{Simonyan15}.
	
	\textbf{Face Identity Loss.} It regularizes the cropped face from the synthesized target image $\hat{I}_t$ to be similar to that from the image of ground truth $I_r$, which pushes the generator to preserve the face identity. It is shown as follows: \vspace{-2mm}
	\begin{equation*}
	\begin{aligned}
	L_f=\|g(\hat{I}_t) - g(I_r)\|_1.
	\end{aligned}
	\label{equ:face}
	\vspace{-2mm}
	\end{equation*}
	Here, $g$ is a pre-trained SphereFaceNet~\cite{cvprLiuWYLRS17}.
	
	\textbf{Adversarial Loss.} It pushes the distribution of synthesized images to the distribution of real images. As shown in following, we use $LSGAN_{-110}$~\cite{lsgan_gp} loss in a way like PatchGAN for the generated target image $\hat{I}_t$. The discriminator $D$ regularizes $\hat{I}_t$ to be more realistic-looking. We use conditioned discriminator, and it takes generated images and the correspondence map $C_t$ (6 channels) as inputs. \vspace{-3mm}
	\begin{equation*}
	\begin{aligned}
	L^G_{adv}&= \sum D(\hat{I}_t, C_t)^2. \\
	\end{aligned}
	\label{equ:adv}
	\vspace{-2mm}
	\end{equation*}
	
	\textbf{Attention Regularization Loss.} It regularizes the attention map $A$ to be smooth and to prevent them from saturating. Considering that there is no ground truth of attention map $A$, as well as color map $P$, they are learned from the resulting gradients of above losses. However, the attention masks can easily saturate to 1 which prevents the generator from working. To alleviate this situation, we regularize the mask to be closer to silhouettes $S$ rendered by 3D body mesh. Since the silhouettes is a rough map and it contains the body mask without clothes and hair,
	we also perform a Total Variation Regularization over A like~\cite{ganimation}, to compensate the shortcomings of silhouettes, and further to enforce smooth spatial color when combining the pixel from the predicted background $\hat{I}_{bg}$ and the color map $P$. It is shown as follows:\vspace{-3mm}
	\begin{equation*}
	\small
	\begin{aligned}
	L_a&= \|A_s-S_s\|^2_2 + \|A_t-S_t\|^2_2 + TV(A_s) + TV(A_t) \\
	TV(A)&=\sum_{i,j}[A(i,j)-A(i-1,j)]^2 + [A(i,j)-A(i,j-1)]^2.
	\end{aligned}
	\label{equ:att}
	\vspace{-4mm}
	\end{equation*}
	For generator, the full objective function is shown in the following, and
	$\lambda_p, \lambda_f$ and $\lambda_a$ are the weights of perceptual, face identity and attention losses. \vspace{-3mm}
	\begin{equation*}
	\begin{aligned}
	L^G = \lambda_p L_p + \lambda_f L_f + \lambda_a L_a + L^G_{adv}.
	\end{aligned}
	\label{equ:l_g}
	\vspace{-3mm}
	\end{equation*}
	For discriminator, the full objective function is \vspace{-2mm}
	\begin{equation*}
	\begin{aligned}
	L^D &= \sum[D(\hat{I}_t, C_t) + 1]^2 + \sum [D(I_r, C_t) - 1]^2.
	\end{aligned}
	\label{equ:l_d}
	\end{equation*}

	\begin{figure*}[t]
		\centering
		\includegraphics[width=0.9\linewidth]{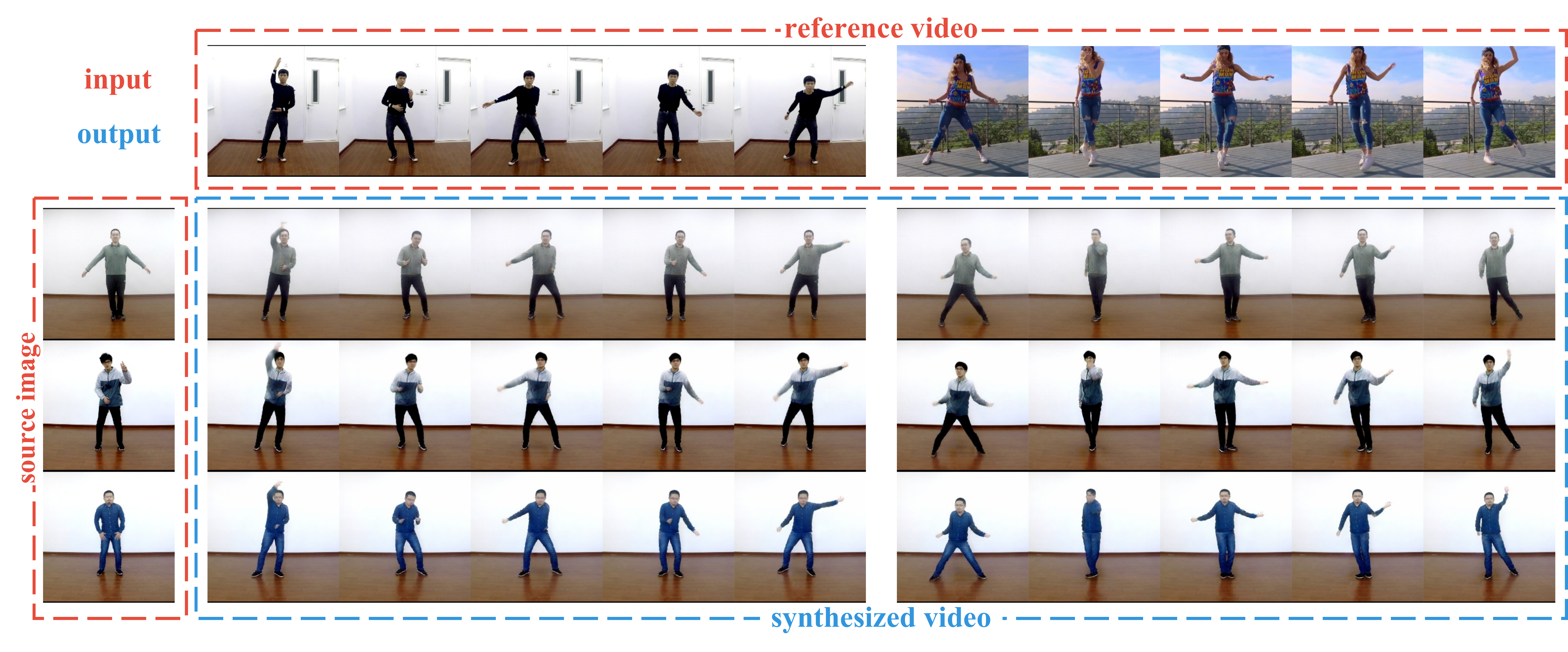}
		\vspace{-3mm}
		\caption{Examples of motion imitation from our proposed methods on the iPER dataset (zoom-in for the best of view). Our method could produce high-fidelity images that preserve the face identity, shape consistency and clothes details of source, even there are occlusions in source images such as the middle and bottom rows. We recommend accessing the supplementary material for more results in videos.}
		\vspace{-4mm}
		\label{fig:imitation} 
	\end{figure*}
	\begin{figure*}[h]
		\centering
		\includegraphics[width=0.9\linewidth]{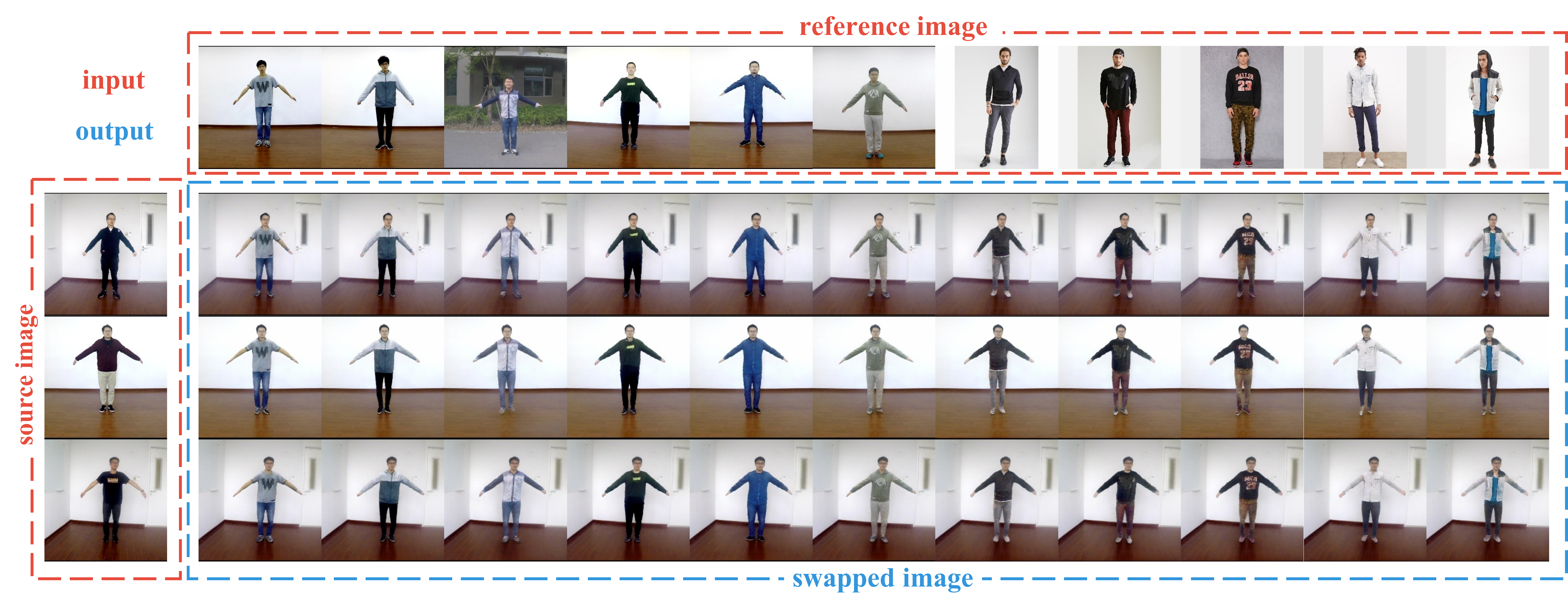}
		\vspace{-3mm}
		\caption{Examples of our method of human appearance transfer in the testing set of iPER (zoom-in for the best of view). Our method could produce high-fidelity and decent images that preserve the face identity and shape consistency of the source image, and keep the clothes details of reference image. We recommend accessing the supplementary material for more results.}
		\label{fig:exam_app} 
		\vspace{-5mm}
	\end{figure*}
	
	\vspace{-2mm}
	\subsection{Inference}
	Once trained model on the task of motion imitation, it can be applied to other tasks in inference. The difference lies in the transformation flow computation, due to the different conditions of various tasks. The remaining modules, Body Mesh Recovery and Liquid Warping GAN modules are all the same. Followings are the details of each task of Flow Composition module in testing phase.
	
	\textbf{Motion Imitation.} We firstly copy the value of pose parameters of reference $\theta_r$ into that of source, and get synthetic parameters of SMPL, as well as the 3D mesh, $M_t = M(\theta_r, \beta_s)$. Next, we render a correspondence map of source mesh $M_s$ and that of synthetic mesh $M_t$ under the camera view of $K_s$. Here, we denote the source and synthetic correspondence map as $C_s$ and $C_t$, respectively. Then, we project vertices of source into 2D image space by weak-perspective camera, $v_s=Proj(V_s, K_s)$. Next, we calculate the barycentric coordinates of each mesh face, and have $f_s \in \mathbb{R}^{N_f \times 2} $. Finally, we calculate the transformation flow $T\in\mathbb{R}^{H\times W\times 2}$ by matching the correspondences between source correspondence map with its mesh face coordinates $f_s$ and synthetic correspondence map. This procedure is shown in Fig.~\ref{fig:inference} (a).
	
	\textbf{Novel View Synthesis.} Given a new camera view, in terms of rotation $R$ and translation $t$. We firstly calculate the 3D mesh under the novel view, $M_t = M_sR + t$. The flowing operations are similar to motion imitation. We render a correspondence map of source mesh $M_s$ and that of novel mesh $M_t$ under the weak-perspective camera $K_s$ and calculate the transformation flow $T\in\mathbb{R}^{H\times W\times 2}$ in the end. This is illustrated in Fig.~\ref{fig:inference} (b).
	
	\textbf{Appearance Transfer.} It needs to ``copy'' the clothes of torso or body from the reference image while keeping the head (face, eye, hair and so on) identity of source. We split the transformation flow $T$ into two sub-transformation flow, source flow $T_1$ and referent flow $T_2$. Denoting head mesh as $M^{h} = (V^{h}, F^{h})$ and body mesh as $M^{b} = (V^{b}, F^{b})$. Here, $M = M^{h} \cup M^{b}$.  For $T_1$, We firstly project the head mesh $M^{h}_s$ of source into image space, and thereby obtain the silhouettes, $S^{h}_s$. Then, we create a mesh grid, $G\in\mathbb{R}^{H\times W\times 2}$. Then, we mask $G$ by $S^{h}$, and derive $T_1 = G * S^{h}$. Here, $*$ represents element-wise multiplication. For $T_2$, it is similar to motion imitation. We render the correspondence map of source body $M^{b}_s$ and that of reference $M^{b}_t$, denoting as $C^{b}_s$ and $C^{b}_t$, respectively. Finally, we calculate the transformation flow $T_2$ based on the correspondences between $C^{b}_s$ and $C^{b}_t$. We illustrate it in Fig.~\ref{fig:inference} (c).
	
	\begin{figure*}
		\centering
		\vspace{-2mm}
		\includegraphics[width=0.9\linewidth]{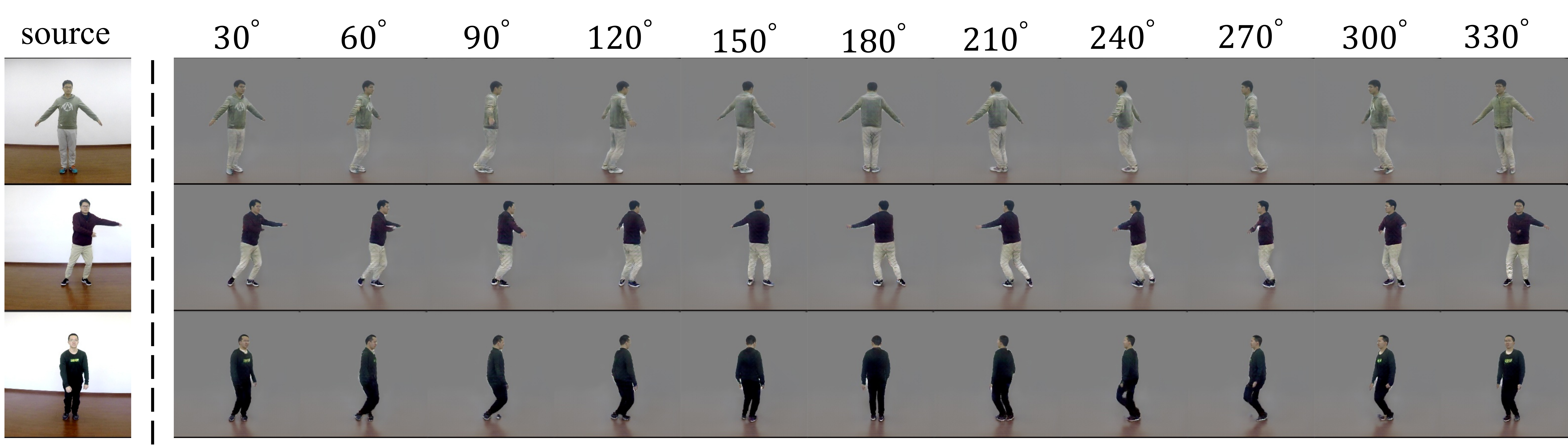}
		\vspace{-2mm}
		\caption{Examples of novel view synthesis from our method on the iPER dataset (zoom-in for the best of view). Our method could generate realistic-looking results under different camera views, and it is capable to preserve the source information, even in the self-occlusion case, such as the middle and bottom rows.}
		\label{fig:novel} 
		\vspace{-4mm}
	\end{figure*}
	
	\section{Experiments}
	\textbf{Dataset}. To evaluate the performances of our proposed method of motion imitation, appearance transfer and novel view synthesis, we build a new dataset with diverse styles of clothes, named as Impersonator (iPER) dataset. There are 30 subjects of different conditions of shape, height and gender. Each subject wears different clothes and performs an A-pose video and a video with random actions. Some subjects might wear multiple clothes, and there are 103 clothes in total.
	The whole dataset contains 206 video sequences with 241,564 frames. We split it into training/testing set at the ratio of 8:2 according to the different clothes.
	
	\textbf{Implementation Details.} To train the network, all images are normalized to [-1, 1] and resized to $256\times256$.  We randomly sample a pair of images from each video. The mini-batch size is 4 in our experiments. $\lambda_p, \lambda_f$ and $\lambda_a$ are set to 10.0, 5.0 and 1.0, respectively. Adam~\cite{Kingma2015AdamAM} is used for parameter optimization of both generator and discriminator.
	
	\subsection{Evaluation of Human Motion Imitation.}
	\vspace{-2mm}
	\textbf{Evaluation Metrics.} We propose an evaluation protocol of testing set of iPER dataset and it is able to indicate the performance of different methods in terms of different aspects. The details are listed in followings:
	1) In each video, we select three images as source images (frontal, sideway and occlusive) with different degrees of occlusion. The frontal image contains the most information, while the sideway will drop out some information, and occlusive image will introduce ambiguity. 2) For each source image, we perform self-imitation that actors imitate actions from themselves. SSIM~\cite{ssimWangBSS04} and Learned Perceptual Similarity (LPIPS)~\cite{zhang2018perceptual} are the evaluation metrics in self-imitation setting. 3) Besides, we also conduct cross-imitation that actors imitate actions from others. We use Inception Score (IS)~\cite{IS_nips2016} and Fr\'{e}chet Distance on a pre-trained person-reid model~\cite{pcb2018eccv}, named as FReID, to evaluate the quality of generated images.
	\begin{table}[t]
		\small
		\begin{center}
			\caption{Results of motion imitation by different methods on iPER dataset. $\uparrow$ means the larger is better, and $\downarrow$ represents the smaller is better. A higher SSIM may not mean a better quality of image~\cite{zhang2018perceptual}.}
			\label{table:all_method}
			\begin{tabular}{c|c|c|c|c}
				\hline
				\multicolumn{1}{c|}{\multirow{2}{*}{}} & \multicolumn{2}{c|}{Self-Imitation}    & \multicolumn{2}{c}{Cross-Imitation}               \\ \cline{2-5} 
				\multicolumn{1}{c|}{}                  & \multicolumn{1}{p{1.0cm}<{\centering}|}{SSIM$\uparrow$}  & \multicolumn{1}{p{1.2cm}<{\centering}|}{LPIPS\footnotemark{}  $\downarrow$} & \multicolumn{1}{p{0.8cm}<{\centering}|}{IS$\uparrow$}    & \multicolumn{1}{p{0.8cm}<{\centering}}{FReID$\downarrow$} \\ \hline
				
				PG2~\cite{pG2017nips} &\textbf{0.854} & 0.135 & 3.242 & 0.353 \\  
				
				SHUP~\cite{posewarp2018} & 0.832 & 0.099 & 3.371 & 0.324 \\  
				
				DSC~\cite{DSC2018} & 0.829 & 0.129 & 3.321 & 0.342 \\ \hline
				
				$W_C$ & 0.821 & 0.128 & 3.213 & 0.341 \\  
				
				$W_T$ & 0.822 & 0.113 & 3.353 & 0.347 \\  
				
				$W_F$ & 0.830 & 0.103 & 3.358 & 0.325 \\	
				\textbf{Ours-$W_{LWB}$}& 0.840 & \textbf{0.087} & \textbf{3.419} & \textbf{0.317} \\ \hline                          
			\end{tabular}
		\end{center}
		\vspace{-5mm}
	\end{table}
	\footnotetext{The results of LPIPS (Learned Perceptual Image Patch Similarity) reported in Table \ref{table:all_method} of our previous paper are slightly different from the original definition~\cite{zhang2018perceptual}. The original definition is the $distance\;score$ between two images, while the previous version uses the $1 - distance\;score$ as the similarity metric. To make it consistent with the original definition, here we update the results with the original definition~\cite{zhang2018perceptual} }
	
	\textbf{Comparison with Other Methods.}
	We compare the performance of our method with that of existing methods, including PG2~\cite{pG2017nips}, SHUP~\cite{posewarp2018} and DSC~\cite{DSC2018}. We train all these methods on iPER dataset, and the evaluation protocol mentioned above is applied to these methods. The results are reported in Table 1. It can be seen that our method outperforms other methods.
	In addition, we also analyze the generated images and make comparisons between ours and above methods. From Fig.~\ref{fig:comparison}, we find that 1) the 2D pose-guided methods, including PG2~\cite{pG2017nips}, SHUP~\cite{posewarp2018} and DSC~\cite{DSC2018}, change the body shape of source. For example, in the $3^{rd}$ row of Fig.~\ref{fig:comparison}, a tall person imitates motion from a short person and these methods change the height of source body. However, our method is capable to keep the body shape unchanged. 2) When source image exhibits self-occlusion, such as invisible face in the $1^{st}$ row of Fig.~\ref{fig:comparison}, our method could generate more realistic-looking content of the ambiguous and invisible parts. 3) Our method is more powerful in terms of preserving source identity, such as the face identity and cloth details of source than other methods, as shown in the $2^{nd}$ and $3^{rd}$ row of Fig.~\ref{fig:comparison}. 4) Our method also produces high-fidelity images in the cross-imitation setting (imitating actions from others) and we illustrate it in Fig.~\ref{fig:imitation}.
	
	\textbf{Ablation Study.}
	To verify the impact of our proposed Liquid Warping Block (LWB), we design three baselines with aforementioned ways to propagate the source information, including early concatenation, texture warping and feature warping. All modules and loss functions are the same except the propagating strategies among our method and other baselines. Here, we denote early concatenation, texture warping, feature warping, and our proposed LWB as $W_C$, $W_T$, $W_F$ and $W_{LWB}$. We train all these under the same setting on the iPER dataset, then evaluate their performances on motion imitation. From Table 1, we can see that our proposed LWB is better than other baselines. More details are provided in supplementary materials.
	\vspace{-1mm}
	\subsection{Results of Human Appearance Transfer.}
	\vspace{-1mm}
	It is worth emphasizing that once model has been trained, it is able to directly to be applied in three tasks, including motion imitation, appearance transfer and novel view synthesis. We randomly pick some examples displayed in Fig.~\ref{fig:exam_app}. The face identity and clothes details, in terms of texture, color and style, are preserved well by our method. It demonstrates that our method can achieve decent results in appearance transfer, even when the reference image comes from Internet and is out of the domain of iPER dataset, such as the last five columns in Fig.~\ref{fig:exam_app}.

	\vspace{-1mm}
	\subsection{Results of Human Novel View Synthesis.}
	\vspace{-1mm}
	We randomly sample source images from the testing set of iPER, and change the views from $30^{\circ}$ to $330^{\circ}$. The results are illustrated in Fig.~\ref{fig:novel}. Our method is capable to predict reasonable content of invisible parts when switching to other views and keep the source information, in terms of face identity and clothes details, even in the self-occlusion case, such as the middle and bottom rows in Fig.~\ref{fig:novel}.
	
	
	\vspace{-1mm}
	\section{Conclusion}
	\vspace{-1mm}
	We propose a unified framework to handle human motion imitation, appearance transfer, and novel view synthesis. It employs a body recovery module to estimate the 3D body mesh which is more powerful than 2D Pose. Furthermore, in order to preserve the source information, we design a novel warping strategy,  Liquid Warping Block (LWB), which propagates the source information in both image and feature spaces, and supports a more flexible warping from multiple sources. Extensive experiments show that our framework outperforms others and produce decent results.
	

\begin{thebibliography}{10}\itemsep=-1pt

\bibitem{posewarp2018}
Guha Balakrishnan, Amy Zhao, Adrian~V. Dalca, Frédo Durand, and John Guttag.
\newblock Synthesizing images of humans in unseen poses.
\newblock In {\em The IEEE Conference on Computer Vision and Pattern
  Recognition (CVPR)}, June 2018.

\bibitem{SMPLify}
Federica Bogo, Angjoo Kanazawa, Christoph Lassner, Peter Gehler, Javier Romero,
  and Michael~J Black.
\newblock Keep it smpl: Automatic estimation of 3d human pose and shape from a
  single image.
\newblock In {\em European Conference on Computer Vision}, pages 561--578.
  Springer, 2016.

\bibitem{chan2018everybody}
Caroline Chan, Shiry Ginosar, Tinghui Zhou, and Alexei~A Efros.
\newblock Everybody dance now.
\newblock {\em arXiv preprint arXiv:1808.07371}, 2018.

\bibitem{softgate18}
Haoye Dong, Xiaodan Liang, Ke Gong, Hanjiang Lai, Jia Zhu, and Jian Yin.
\newblock Soft-gated warping-gan for pose-guided person image synthesis.
\newblock In {\em Advances in Neural Information Processing Systems 31: Annual
  Conference on Neural Information Processing Systems 2018, NeurIPS 2018, 3-8
  December 2018, Montr{\'{e}}al, Canada.}, pages 472--482, 2018.

\bibitem{vunet2018}
Patrick Esser, Ekaterina Sutter, and Bj{\"{o}}rn Ommer.
\newblock A variational u-net for conditional appearance and shape generation.
\newblock In {\em {IEEE} Conference on Computer Vision and Pattern
  Recognition}, pages 8857--8866, 2018.

\bibitem{gan2014}
Ian Goodfellow, Jean Pouget-Abadie, Mehdi Mirza, Bing Xu, David Warde-Farley,
  Sherjil Ozair, Aaron Courville, and Yoshua Bengio.
\newblock Generative adversarial nets.
\newblock In Z. Ghahramani, M. Welling, C. Cortes, N.~D. Lawrence, and K.~Q.
  Weinberger, editors, {\em Advances in Neural Information Processing Systems
  27}, pages 2672--2680. Curran Associates, Inc., 2014.

\bibitem{resnetHe16}
Kaiming He, Xiangyu Zhang, Shaoqing Ren, and Jian Sun.
\newblock Deep residual learning for image recognition.
\newblock In {\em 2016 {IEEE} Conference on Computer Vision and Pattern
  Recognition, {CVPR} 2016, Las Vegas, NV, USA, June 27-30, 2016}, pages
  770--778, 2016.

\bibitem{resnetv2He16}
Kaiming He, Xiangyu Zhang, Shaoqing Ren, and Jian Sun.
\newblock Identity mappings in deep residual networks.
\newblock In {\em Computer Vision - {ECCV} 2016 - 14th European Conference,
  Amsterdam, The Netherlands, October 11-14, 2016, Proceedings, Part {IV}},
  pages 630--645, 2016.

\bibitem{Isola2017ImagetoImageTW}
Phillip Isola, Jun-Yan Zhu, Tinghui Zhou, and Alexei~A. Efros.
\newblock Image-to-image translation with conditional adversarial networks.
\newblock {\em 2017 IEEE Conference on Computer Vision and Pattern Recognition
  (CVPR)}, pages 5967--5976, 2017.

\bibitem{STN2015}
Max Jaderberg, Karen Simonyan, Andrew Zisserman, and Koray Kavukcuoglu.
\newblock Spatial transformer networks.
\newblock In {\em Advances in Neural Information Processing Systems 28: Annual
  Conference on Neural Information Processing Systems 2015, December 7-12,
  2015, Montreal, Quebec, Canada}, pages 2017--2025, 2015.

\bibitem{eccvJohnsonAF16}
Justin Johnson, Alexandre Alahi, and Li Fei{-}Fei.
\newblock Perceptual losses for real-time style transfer and super-resolution.
\newblock In {\em Computer Vision - {ECCV} 2016 - 14th European Conference,
  Amsterdam, The Netherlands, October 11-14, 2016, Proceedings, Part {II}},
  pages 694--711, 2016.

\bibitem{HMR}
Angjoo Kanazawa, Michael~J Black, David~W Jacobs, and Jitendra Malik.
\newblock End-to-end recovery of human shape and pose.
\newblock In {\em The IEEE Conference on Computer Vision and Pattern
  Recognition (CVPR)}, 2018.

\bibitem{cvprKatoUH18}
Hiroharu Kato, Yoshitaka Ushiku, and Tatsuya Harada.
\newblock Neural 3d mesh renderer.
\newblock In {\em 2018 {IEEE} Conference on Computer Vision and Pattern
  Recognition, {CVPR} 2018, Salt Lake City, UT, USA, June 18-22, 2018}, pages
  3907--3916, 2018.

\bibitem{Kingma2015AdamAM}
Diederik~P. Kingma and Jimmy Ba.
\newblock Adam: A method for stochastic optimization.
\newblock In {\em International Conference on Learning Representations}, volume
  abs/1412.6980, 2015.

\bibitem{mvLeroyFB17}
Vincent Leroy, Jean{-}S{\'{e}}bastien Franco, and Edmond Boyer.
\newblock Multi-view dynamic shape refinement using local temporal integration.
\newblock In {\em {IEEE} International Conference on Computer Vision, {ICCV}
  2017, Venice, Italy, October 22-29, 2017}, pages 3113--3122, 2003.

\bibitem{Liu2018Neural}
Lingjie Liu, Weipeng Xu, Michael Zollhoefer, Hyeongwoo Kim, Florian Bernard,
  Marc Habermann, Wenping Wang, and Christian Theobalt.
\newblock Neural rendering and reenactment of human actor videos.
\newblock {\em ACM Transactions on Graphics 2019 (TOG)}, 2019.

\bibitem{cvprLiuWYLRS17}
Weiyang Liu, Yandong Wen, Zhiding Yu, Ming Li, Bhiksha Raj, and Le Song.
\newblock Sphereface: Deep hypersphere embedding for face recognition.
\newblock In {\em 2017 {IEEE} Conference on Computer Vision and Pattern
  Recognition, {CVPR} 2017, Honolulu, HI, USA, July 21-26, 2017}, pages
  6738--6746, 2017.

\bibitem{SMPL:2015}
Matthew Loper, Naureen Mahmood, Javier Romero, Gerard Pons-Moll, and Michael~J.
  Black.
\newblock {SMPL}: A skinned multi-person linear model.
\newblock {\em ACM Trans. Graphics (Proc. SIGGRAPH Asia)}, 34(6):248:1--248:16,
  oct 2015.

\bibitem{pG2017nips}
Liqian Ma, Xu Jia, Qianru Sun, Bernt Schiele, Tinne Tuytelaars, and Luc
  Van~Gool.
\newblock Pose guided person image generation.
\newblock In {\em Advances in Neural Information Processing Systems}, pages
  405--415, 2017.

\bibitem{ma2018disentangled}
Liqian Ma, Qianru Sun, Stamatios Georgoulis, Luc Van~Gool, Bernt Schiele, and
  Mario Fritz.
\newblock Disentangled person image generation.
\newblock In {\em {IEEE} Conference on Computer Vision and Pattern
  Recognition}, 2018.

\bibitem{lsgan_gp}
Xudong Mao, Qing Li, Haoran Xie, Raymond Y.~K. Lau, Zhen Wang, and Stephen~Paul
  Smolley.
\newblock On the effectiveness of least squares generative adversarial
  networks.
\newblock {\em CoRR}, abs/1712.06391, 2017.

\bibitem{DensePoseTransfer}
Natalia Neverova, R{\i}za~Alp G{\"u}ler, and Iasonas Kokkinos.
\newblock Dense pose transfer.
\newblock In {\em European Conference on Computer Vision (ECCV)}, 2018.

\bibitem{Park_2017_CVPR}
Eunbyung Park, Jimei Yang, Ersin Yumer, Duygu Ceylan, and Alexander~C. Berg.
\newblock Transformation-grounded image generation network for novel 3d view
  synthesis.
\newblock In {\em The IEEE Conference on Computer Vision and Pattern
  Recognition (CVPR)}, July 2017.

\bibitem{ponsmollSIGGRAPH17clothcap}
Gerard Pons{-}Moll, Sergi Pujades, Sonny Hu, and Michael~J. Black.
\newblock Clothcap: seamless 4d clothing capture and retargeting.
\newblock {\em {ACM} Trans. Graph.}, 36(4):73:1--73:15, 2017.

\bibitem{ganimation}
Albert Pumarola, Antonio Agudo, Aleix~M. Martinez, Alberto Sanfeliu, and
  Francesc Moreno{-}Noguer.
\newblock Ganimation: Anatomically-aware facial animation from a single image.
\newblock In {\em Computer Vision - {ECCV} 2018 - 15th European Conference,
  Munich, Germany, September 8-14, 2018, Proceedings, Part {X}}, pages
  835--851, 2018.

\bibitem{swapnet2018}
Amit Raj, Patsorn Sangkloy, Huiwen Chang, James Hays, Duygu Ceylan, and Jingwan
  Lu.
\newblock Swapnet: Image based garment transfer.
\newblock In {\em Computer Vision - {ECCV} 2018 - 15th European Conference,
  Munich, Germany, September 8-14, 2018, Proceedings, Part {XII}}, pages
  679--695, 2018.

\bibitem{DensePose}
Iasonas~Kokkinos R{\i}za Alp~G{\"u}ler, Natalia~Neverova.
\newblock Densepose: Dense human pose estimation in the wild.
\newblock In {\em The IEEE Conference on Computer Vision and Pattern
  Recognition (CVPR)}, 2018.

\bibitem{unet2015}
Olaf Ronneberger, Philipp Fischer, and Thomas Brox.
\newblock U-net: Convolutional networks for biomedical image segmentation.
\newblock In {\em Medical Image Computing and Computer-Assisted Intervention -
  {MICCAI} 2015 - 18th International Conference Munich, Germany, October 5 - 9,
  2015, Proceedings, Part {III}}, pages 234--241, 2015.

\bibitem{IS_nips2016}
Tim Salimans, Ian~J. Goodfellow, Wojciech Zaremba, Vicki Cheung, Alec Radford,
  and Xi Chen.
\newblock Improved techniques for training gans.
\newblock In {\em Advances in Neural Information Processing Systems 29: Annual
  Conference on Neural Information Processing Systems 2016, December 5-10,
  2016, Barcelona, Spain}, pages 2226--2234, 2016.

\bibitem{Si_2018_CVPR}
Chenyang Si, Wei Wang, Liang Wang, and Tieniu Tan.
\newblock Multistage adversarial losses for pose-based human image synthesis.
\newblock In {\em The IEEE Conference on Computer Vision and Pattern
  Recognition (CVPR)}, June 2018.

\bibitem{DSC2018}
Aliaksandr Siarohin, Enver Sangineto, Stéphane Lathuilière, and Nicu Sebe.
\newblock Deformable gans for pose-based human image generation.
\newblock In {\em The IEEE Conference on Computer Vision and Pattern
  Recognition (CVPR)}, June 2018.

\bibitem{Simonyan15}
Karen Simonyan and Andrew Zisserman.
\newblock Very deep convolutional networks for large-scale image recognition.
\newblock In {\em 3rd International Conference on Learning Representations,
  {ICLR} 2015, San Diego, CA, USA}, May 2015.

\bibitem{pcb2018eccv}
Yifan Sun, Liang Zheng, Yi Yang, Qi Tian, and Shengjin Wang.
\newblock Beyond part models: Person retrieval with refined part pooling (and
  {A} strong convolutional baseline).
\newblock In {\em Computer Vision - {ECCV} 2018 - 15th European Conference,
  Munich, Germany, September 8-14, 2018, Proceedings, Part {IV}}, pages
  501--518, 2018.

\bibitem{wangVID2VID}
Ting{-}Chun Wang, Ming{-}Yu Liu, Jun{-}Yan Zhu, Nikolai Yakovenko, Andrew Tao,
  Jan Kautz, and Bryan Catanzaro.
\newblock Video-to-video synthesis.
\newblock In {\em Advances in Neural Information Processing Systems 31: Annual
  Conference on Neural Information Processing Systems 2018, NeurIPS 2018, 3-8
  December 2018, Montr{\'{e}}al, Canada.}, pages 1152--1164, 20148.

\bibitem{Wang_2018_CVPR}
Ting-Chun Wang, Ming-Yu Liu, Jun-Yan Zhu, Andrew Tao, Jan Kautz, and Bryan
  Catanzaro.
\newblock High-resolution image synthesis and semantic manipulation with
  conditional gans.
\newblock In {\em The IEEE Conference on Computer Vision and Pattern
  Recognition (CVPR)}, June 2018.

\bibitem{ssimWangBSS04}
Zhou Wang, Alan~C. Bovik, Hamid~R. Sheikh, and Eero~P. Simoncelli.
\newblock Image quality assessment: from error visibility to structural
  similarity.
\newblock {\em {IEEE} Trans. Image Processing}, 13(4):600--612, 2004.

\bibitem{HAT_2018_CVPR}
Mihai Zanfir, Alin-Ionut Popa, Andrei Zanfir, and Cristian Sminchisescu.
\newblock Human appearance transfer.
\newblock In {\em The IEEE Conference on Computer Vision and Pattern
  Recognition (CVPR)}, June 2018.

\bibitem{3dScanZhang17}
Chao Zhang, Sergi Pujades, Michael~J. Black, and Gerard Pons{-}Moll.
\newblock Detailed, accurate, human shape estimation from clothed 3d scan
  sequences.
\newblock In {\em 2017 {IEEE} Conference on Computer Vision and Pattern
  Recognition, {CVPR} 2017, Honolulu, HI, USA, July 21-26, 2017}, pages
  5484--5493, 2017.

\bibitem{zhang2018perceptual}
Richard Zhang, Phillip Isola, Alexei~A Efros, Eli Shechtman, and Oliver Wang.
\newblock The unreasonable effectiveness of deep features as a perceptual
  metric.
\newblock In {\em The IEEE Conference on Computer Vision and Pattern
  Recognition (CVPR)}, 2018.

\bibitem{Zhao0C0JF18}
Bo Zhao, Xiao Wu, Zhi{-}Qi Cheng, Hao Liu, Zequn Jie, and Jiashi Feng.
\newblock Multi-view image generation from a single-view.
\newblock In {\em 2018 {ACM} Multimedia Conference on Multimedia Conference,
  {MM} 2018, Seoul, Republic of Korea, October 22-26, 2018}, pages 383--391,
  2018.

\bibitem{ZhouTSME16}
Tinghui Zhou, Shubham Tulsiani, Weilun Sun, Jitendra Malik, and Alexei~A.
  Efros.
\newblock View synthesis by appearance flow.
\newblock In {\em Computer Vision - {ECCV} 2016 - 14th European Conference,
  Amsterdam, The Netherlands, October 11-14, 2016, Proceedings, Part {IV}},
  pages 286--301, 2016.

\bibitem{Zhu_2018_CVPR}
Hao Zhu, Hao Su, Peng Wang, Xun Cao, and Ruigang Yang.
\newblock View extrapolation of human body from a single image.
\newblock In {\em The IEEE Conference on Computer Vision and Pattern
  Recognition (CVPR)}, June 2018.

\end{thebibliography}

\end{document}